# An Algorithm for Deciding if a Set of Observed Independencies Has a Causal Explanation

Thomas Verma
Northrop Corporation
One Research Park
Palos Verdes, CA 90274
verma@nrtc.northrop.com

Judea Pearl
Cognitive Systems Laboratory
Computer Science Department
University of California
Los Angeles, CA 90024
judea@cs.ucla.edu

## Abstract

In a previous paper [Pearl and Verma, 1991] we presented an algorithm for extracting causal influences from independence information, where a causal influence was defined as the existence of a directed arc in all minimal causal models consistent with the data. In this paper we address the question of deciding whether there exists a causal model that explains ALL the observed dependencies and independencies. Formally, given a list M of conditional independence statements, it is required to decide whether there exists a directed acyclic graph (dag) D that is perfectly consistent with M, namely, every statement in M, and no other, is reflected via d-separation in D. We present and analyze an effective algorithm that tests for the existence of such a dag, and produces one, if it exists.

## 1 Introduction

Directed acyclic graphs (dags) have been widely used for modeling statistical data. Starting with the pioneering work of Sewal Wright [Wright, 1921] who introduced path analysis to statistics, through the more recent development of Bayesian networks and influence diagrams, dag structures have served primarily for encoding causal influences between variables as well as between actions and variables.

Even statisticians who usually treat causality with extreme caution, have found the structure of dags to be an advantageous model for explanatory purposes. N. Wermuth, for example, mentions several such advantages [Wermuth, 1991]. First, the dag describes a stepwise stochastic process by which the data *could have been* generated and in this sense it may even "prove the basis for developing causal explanations" [Cox, 1992]. Second, each parameter in the dag has a well understood meaning since it is a conditional probability, i.e., it measures the probability of the response variable given a particular configuration of the explanatory (parents) variables and all other variables being unspecified. Third, the task of estimating the parameters in the dag model can be decomposed into a sequence of local estimation analyses, each involving a variable and its parent set in the dag. Fourth, general results are available for reading all implied independencies directly off the dag [Verma, 1986], [Pearl, 1988], [Lauritzen et al., 1990] and for deciding from the topology of two given dags whether they are equivalent, i.e., whether they specify the same set of independence-restrictions on the joint distribution [Frydenberg, 1990], [Verma and Pearl, 1990], and whether one dag specifies more restrictions than the other [Pearl et al., 1989][1].

This paper adds a fifth advantage to the list above. It presents an algorithm which decides for an arbitrary list of conditional independence statements whether it defines a dag and, if it does, a corresponding dag is drawn. The algorithm we present has its basis in the "Inferred-Causation" (IC) algorithm described in [Pearl and Verma, 1991] and in Lemmas 1 and 2 of [Verma and Pearl, 1990]. Whereas in [Pearl and Verma, 1991] we were interested in detecting local relationships that we called "genuine causal influences", we now consider an entire dag as one unit which ought to fit the data at hand.

### 1.1 Problem

Given a list $M$ of conditional independence statements ranging over a set of variables $U$ it is required to decide whether there exists a directed acyclic graph (dag) $D$ that is consistent with $M$.

Our analysis will focus on lists that are closed under the graphoid axioms (see Appendix for definition). Section 5 will discuss possible extensions to lists which are not closed.

---

### 1.2  Definitions

A *dependency model* is a list of conditional independence statements of the form $I(A, B|C)$, where $A$, $B$ and $C$ are disjoint subsets of some set of variables $U$. A dag $D$ is *consistent* with a dependency model $M$ if every statement in $M$ and no statement outside $M$ follows from the topology of $D$. In this case, $M$ is said to be *dag-isomorphic*. A statement $I$ follows from the topology of a dag $D$, if $I$ holds in every probability distribution $P$ that is compatible with $D^2$ can be decomposed into a product of conditional probabilities $P(a|\bar{\pi}(a))$, over all nodes $a \in U$, where $\bar{\pi}(a)$ is a set containing the parents of $a$ in $D$. Finally, a statement $I(A, B|C)$ holds in a probability distribution $P$ iff $P(A|C)P(B|C) = P(AB|C)$.

The following definitions and notation are needed to understand the proposed solution. A *partially directed acyclic graph* (pdag) is a graph which contains both directed and undirected edges, but it does not contain any directed cycles. An *extension* of a pdag $G$, is any fully directed acyclic graph, $D$, which has the same skeleton (underlying undirected edges) as $G$ and the same vee structures as $G$. Three nodes form a *vee structure*, written $\overset{\frown}{abc}$, if $a \rightarrow b \leftarrow c$ and $a$ is not adjacent to $c$. Two nodes are *adjacent*, written $\overline{ab}$, if either $a \rightarrow b$, $a \leftarrow b$ or $a - b$.

### 1.3  Overview

Section 2 details the solution to the problem posed in Section 1.1. It presents an algorithm which consists of the following three phases.

- Phase 1 examines the independence statements in $M$ and tries to construct a pdag, $G$ with the following guarantees:

  1. If $M$ is dag-isomorphic then every extension of $G$ will be consistent with $M$.

  2. If Phase 1 fails to generate a pdag, then $M$ is not dag-isomorphic.

- Phase 2 extends a pdag, $G$, into a dag $D$, if possible.

- Phase 3 verifies if $D$ is consistent with $M$.

If $D$ is found to be consistent with $M$ then $M$ is dag-isomorphic, by definition. If $D$ is found to be inconsistent with $M$ then $M$ is not dag-isomorphic and (by definition) no dag can be consistent with $M$.

Additional improvements to this algorithm and extensions to the problem are discussed in Section 5.

---

[2]Alternatively, such a statement corresponds to a d-separation condition in $D$ [Pearl, 1988].

## 2  The DAG Construction Algorithm

### Phase 1

Generate a pdag $G$, from $M$, if possible.

1. For each pair of variables, $(a, b)$, look through $M$ for a statement of the form $I(a, b|S)$, where $S$ is any set of variables (including $\emptyset$). Construct an undirected graph $G$ where vertices $a$ and $b$ are connected by an edge iff a statement $I(a, b|S)$ is not found in $M$. Mark every pair of non-adjacent nodes in $G$ with the set $S$ found in $M$, call this set $S(a, b)$.

2. For every pair of non-adjacent nodes $a$ and $c$ in $G$, test if there is a node $b$ not in $S(a, c)$ that is adjacent to both $a$ and $c$. If there is such a node then direct the arcs $a \rightarrow b$ and $c \rightarrow b$ unless there already exists a directed path from $b$ to $a$ or from $b$ to $c$, in which case Phase 1 FAILS.

3. If the orientation of Step 2 is completed then Phase 1 SUCCEEDS, and returns a partially directed graph, $G$.

### Phase 2

Extend $G$ into a dag, $D$, if possible.

1. Initially let $C$ be an empty stack and let $D$ equal $G$.

2. While $D$ contains any undirected arcs repeat 2a, 2b and 2c:

   (a) Close $D$ under the following four rules, if possible.

   **Rule 1:** If $a \rightarrow b - c$ and $a$ is not adjacent to $c$ then direct $b \rightarrow c$.

   **Rule 2:** If 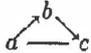 then $a \rightarrow c$.

   **Rule 3:** If 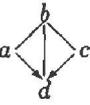 then direct $b \rightarrow d$.

   **Rule 4:** If 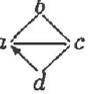 then direct $a \rightarrow b$ and $c \rightarrow b$.

   (b) If the closure was successful, i.e. there are no directed cycles or new vee structures, then:

   - If $D$ still contains any undirected arcs, select one and choose a direction for it, push the arc and a copy of $D$ onto the stack $C$ and continue the while loop (i.e. go back to 2a).
   - If $G$ contains no more undirected arcs, then the while loop is completed, Phase 2 SUCCEEDS, and returns a directed acyclic graph $D$.

(c) If the closure was unsuccessful, then discard the current value of $D$ and pop the most recent copy off of the stack along with the selected arc. Reverse the chosen direction of the arc in $D$ and continue the while loop (i.e. go back to 2a).

## Phase 3

Check if $D$ is consistent with $M$.

1. Test that every statement $I$ in $M$ holds in $D$ (using the d-separation criterion)[3].

2. Pick any total ordering of the nodes which agrees with the directionality of the $D$ and let $U_a$ stand for the set of nodes which precede $a$ in this ordering. For every node $a$ in $D$, test if the statement $I(a, U_a \setminus \bar{\pi}(a) | \bar{\pi}(a))$ is in M.

3. If both tests are confirmed, EXIT with SUCCESS, and return $D$; else, EXIT with FAIL.

## 3    Correctness

### Phase 1

This phase examines $M$ and generates a graph, $G$ subject to the above guarantees, if possible. That is, if $M$ is dag-isomorphic then every extension of $G$ is consistent with $M$. The correctness of Step 1 of this phase follows from Lemma 3.1 [Verma, 1991][Verma and Pearl, 1990] (a detailed proof of which is given in the appendix). This lemma is also the basis for the inference algorithm developed by Spirtes and Glymour [Spirtes and Glymour, 1991].

**Lemma 3.1** *Let $M$ be any dag isomorphic dependency model, a dag $D$ is consistent with $M$ iff the following two conditions hold:*

*1.* $\overline{ab}$ *in $D$ iff $\forall_S$, $I(a, b|S) \notin M$.*

*2.* $\overline{abc}$ *in $D$ iff $\overline{abc}$ and $\neg \overline{ac}$ (using condition 1) in $D$ and $\forall_S$, if $I(a, c|S) \in M$ then $b \notin S$.*

**Corollary 3.2** *Two dags are equivalent iff they share the same set of links and same set of vee structures.*

The *only-if* portion of this lemma guarantees that:

1. If there exists some dag $D^*$ which is consistent with $M$, then any dag $D$ consistent with $M$ must have the same skeleton as $D^*$.

2. Furthermore, every dag $D$, consistent with $M$ must have the same vee structures as $D^*$.

The *if* part guarantees that every dag $D$ which has the same skeleton and vee structures as $D^*$, is consistent with $M$. The first step of Phase 1 attempts to construct this invariant skeleton if $M$ is dag-isomorphic. The arrowheads added in the second step identify the invariant vee structures, again, if $M$ is dag-isomorphic.

Note however, that Step 2 of Phase 1 directs arcs immediately upon finding one set $S$ satisfying condition 2 of the lemma. This decision is correct due to the following lemma:

**Lemma 3.3** *For any dag-isomorphic dependency model $M$ and any three variables $a$, $b$ and $c$ forming a chain $\overline{abc}$, if $\exists_S$ s.t. $I(a, c|S) \in M$ and $b \notin S$ then $\forall'_S$ $I(a, c|S') \in M$ implies $b \notin S'$.*

This lemma permits the use of the first $S$ found to orient the vee structures.

If $M$ is not dag-isomorphic it would be possible for Phase 1 to build a graph that is not a pdag if it weren't for the failure condition in Step 2. The next example illustrates a failure resulting from an application of Phase 1 on a non-dag-isomorphic dependency model.

**Example 3.4** Set $U = \{a, b, c, d\}$ and $M$ be the closure of the set $\{I(a, c|\emptyset), I(a, d|\emptyset), I(b, d|\emptyset)\}$ under symmetry[4].

Step 1 of Phase 1 will construct the skeleton $a-b-c-d$, and $S(a, c) = S(a, d) = S(b, d) = \emptyset$. Since there is a chain $\overline{abc}$ and $b \notin S(a, c)$ Step 2 could direct $a \rightarrow b \leftarrow c$. Similarly since $\overline{bcd}$ and $\neg \overline{bd}$ and $c \notin S(b, d)$, Step 2 could direct $b \rightarrow c \leftarrow d$.

One of the two directions would be assigned first, then upon attempting the second the algorithm would FAIL.

### Phase 2

The task of Phase 2 is to find a whether a pdag, $G$, has any extensions and to find one if such exists. This is a purely graph theoretic task; it does not involve $M$.

To prove that this phase of the construction is correct, it is sufficient to prove that each of the four rules is sound, namely, that the orientation choices dictated by these rules never need to be revoked.

- Rule 1: If $a \rightarrow b - c$ and $a$ is not adjacent to $c$ then direct $b \rightarrow c$.

  Directing $b - c$ as $b \leftarrow c$ would create a new vee structure, $\overline{abc}$, thus if there is a consistent extension it must contain $b \rightarrow c$.

- Rule 2: If 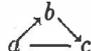 then $a \rightarrow c$.

---

[3]A linear time algorithm for testing d-separation is reported in [Geiger et al., 1990].

[4]Symmetry states that $I(A, B|C)$ iff $I(B, A|C)$. Unless otherwise noted, dependency models are assumed to be closed under symmetry since this is a trivial operation.

Directing $a - c$ as $a \leftarrow c$ would create a directed cycle, $[abca]$, thus if there is a consistent extension it must contain $a \rightarrow c$.

- Rule 3: If 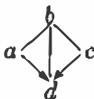 then direct $b \rightarrow d$.

Directing $b - d$ as $b \leftarrow d$ would imply that $a - b$ must be directed as $a \rightarrow b$ or else there would be a directed cycle, $[adba]$. Now if $b - c$ is directed as $b \rightarrow c$ then there is a directed cycle, $[bcdb]$, and if it is directed as $b \leftarrow c$ then there is a new vee structure, $\overrightarrow{abc}$. Thus if there is a consistent extension it must contain $b \rightarrow d$.

- Rule 4: If 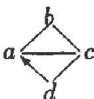 then direct $a \rightarrow b \leftarrow c$.

First, $a - b$ must be directed as $a \rightarrow b$ or there would be a new vee structure, $\overrightarrow{dab}$. If $b - c$ is directed as $b \rightarrow c$ then $c - d$ cannot be directed as $c \rightarrow d$ or there would be a directed cycle, $[cdabc]$. Moreover, $c - d$ cannot be directed as $c \leftarrow d$ or there would be a new vee structure, $\overrightarrow{bcd}$. Thus if there is a consistent extension, then it must contain $a \rightarrow b \leftarrow c$.

Following are two simple examples of pdags which cannot be extended into dags.

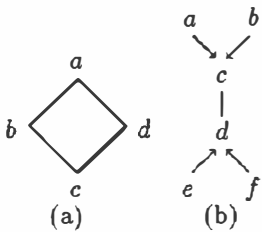

Figure 1: Two pdags which cannot be extended.

**Example 3.5** Consider the graph of Figure 1.a. Initially, no rules apply, so the algorithm would select an arbitrary arc and direct it, without loss of generality assume it directs $a \rightarrow b$. Now Rule 1 will apply twice, directing $b \rightarrow c \rightarrow d$. However a third application to infer $d \rightarrow a$ would produce a directed cycle. It is easy to see that a cycle would result no matter which arc is initially chosen and no matter what initial directionality is assigned. Thus this graph has no dag extension.

**Example 3.6** Consider the graph of Figure 1.b. Any application of Rule 1 to direct the arc $c - d$ would create a new vee structure. Hence this graph as well, has no dag extension.

**Phase 3**

The soundness of Step 1 follows from the definition of consistency; it simply checks if each and every independence statement of $M$ is represented in $D$. The soundness of Step 2, namely that testing only statments of the form $I(a, U_a \setminus \bar{\pi}(a) | \bar{\pi}(a))$ is sufficient follows from the proof of the soundness of d-separation[Verma, 1986].

**Example 3.7** Let $U = a, b, c$ and $M = \{I(a, b|\emptyset), I(a, c|\emptyset), I(b, c|\emptyset)\}$. Phase 1 will produce an empty graph which can trivially be extended into an empty dag. But every independence statement is true in an empty dag, including, e.g. $I(a, b|c)$ which is not in $M$. Thus $M$ is not dag isomorphic.

## 4    Complexity Analysis

Phase 1 can be completed in $O(|M| + |U|^2)$ steps, as follows:

- Start with a complete graph $G$. For each statement, $I(A, B|S)$ in $M$, and for each pair of variables $a \in A$, and $b \in B$ remove the links $a - b$ from $G$ and define $S(a, b) = S$.

- For each node $a$ let $N(a) = \{b | a - b\}$ be the set of neighbors of $a$.

- For each separating set $S(a, b)$ defined above, note that $C(a, b) = N(a) \cup N(b) \setminus S(a, b)$ must be children of $a$ and $b$ so direct $a \rightarrow c \leftarrow b \; \forall c \in C(a, b)$.

Phase 2 may appear to require an exponential amount of time in the worst case due to possible backtracking in Step 2(c). However, we conjecture that if $G$ is extendible, then Rules 1-4 are sufficient to guarantee that no choice will ever need to be revoked. Empirical studies have, so far, confirmed our conjecture. Furthermore, [Verma, 1992] presents an alternative algorithm for Phase 2 based on the maximum cardinality search developed by [Tarjan and Yannakakis, 1984], and which is provably a linear-time algorithm. This algorithm, however, is considerably more complicated and less intuitive than the one presented here.

If the conjecture is correct, it would be possible to replace the backtrack step with a definite failure, in which case the time complexity of this phase would be polynomial, no more than $O(|U|^4 * |E|)$. On the other hand, if it is not correct, the complexity could be exponential in $|E|$.

Phase 3 can be completed in $O(|M| * |E| + |M| * |U|)$ steps.

## 5    Extensions and Improvements

In general, the set of all independence statements which hold for a given domain will grow exponentially as the number of variables grows. Thus it might be impractical to specify $M$ by explicit enumeration of its I-statements. In such cases it may be desirable, instead, to specify a basis, $L$, such that $M$ is the logical closure of $L$, (i.e. $M = CL(L)$), relative to some semantics, (e.g. the graphoid axioms, correlational graphoids axioms, or even probability theory).

The major difficulty in permitting the dependency model to be specified as the closure of some basis lies in solving the so called *membership problem*. Simply stated, the problem is to decide if a particular statement, $I_0$, is contained in the closure, $M$, of a given list of statements, $L$. In general, membership problems are often undecidable, and of those that are decidable, many are NP-hard. In particular, the membership problems for both graphoids and probabilistic independence are unsolved [Geiger, 1990].

However, in spite of this difficulty, it may still be possible to have an efficient dag construction algorithm, because the queries required are of a special form. The algorithm makes four types of queries to $M$:

1. "Is there any $S$ such that $I(a,b|S) \in CL(L)$?" (Phase 1, Step 1)

2. "Is $b$ in any set $S$ such that $I(a,c|S) \in CL(L)$?" (Phase 1, Step 2)

3. "Is every statement in $CL(L)$ represented in $D$?" (Phase 3, Step 1)

4. "Is every statement represented in $D$ in $CL(L)$?" (Phase 3, Step 2)

In the case that $M$ is assumed to be the graphoid closure of $L$, queries of type 1, 2 and 3 are all manageable. The queries for Phase 1 can both be quickly answered due to the following lemma[5]:

**Lemma 5.1** *If* $\exists_S$ *s.t.* $I(a,b|S) \in CL(L)$ *then* $\exists_{A,B,C}$ *s.t.* $I(aA, bB|C) \in L$

Remark: Note that this simplification is possible due to the special form of these queries, namely that $a$ and $b$ are both singletons and any separating set will suffice.

Type 3 queries pose no particular problem since the axioms of graphoids hold for d-separation. Thus it is enough to check that each statement in $L$ is represented in $D$ to ensure that the every statement in closure of $L$ is represented in $D$.

However, to check that each statement represented in $D$ is contained in $CL(L)$ it is necessary to make the $|U|$ membership queries explicated in Step 2 of Phase 3. Although these statements have a special form, it is yet unclear whether a lemma similar to 5.1 exits to simplify these queries.

---

[5]This lemma follows immediately from the form of the graphoid axioms.

Another possible source for simplification is to note that the dag $D$ being tested in Step 2 of Phase 3 is not an arbitrary dag, but the output of the construction algorithm. While Example 3.7 demonstrates that it is possible for $D$ to contain I-statements which are not in $CL(L)$, it may still be the case that any such I-statements must have either a certain form or some other property that would simplify the membership query.

## Acknowledgement

We would like to thank P. Spirtes and A. Paz for many useful discussions. This work was supported in part by National Science Foundation grant #IRI-88-21444 and State of California MICRO grants #90-126 and #90-127.

## Appendix: Proof of Lemmas

**Definition A.1 (d-separation)** *For any dag $D$, two disjoint sets of nodes, $X$, $Y$ are* **d-separated** *given a third $Z$, written $I_D(X,Y|Z)$, if and only if no path between any node in $X$ and any node in $Y$ is activated by the set $Z$.*

*A* **path** *is* **active** *given a set $Z$ if and only if every head to head node of the path is active given $Z$ and every other node of the path is not in $Z$.*

*A* **node** *is* **active** *given a set $Z$ if and only if there is a directed path from it to some element of $Z$.*

The three equivalent terms $Z$-**active**, "active given $Z$" and "activated by $Z$" are used interchangeably.

**Lemma 3.1** *Let $M$ be any dag isomorphic dependency model, a dag $D$ is consistent with $M$ iff the following two conditions hold:*

1. *$\overline{ab}$ in $D$ iff $\forall_S$, $I(a,b|S) \notin M$.*

2. *$\overrightarrow{abc}$ in $D$ iff $\overleftrightarrow{abc}$ and $\neg \overline{ac}$ in $D$ and $\forall_S$, if $I(a,c|S) \in M$ then $b \notin S$.*

**Proof:** There are three basic parts to the proof, (1) that the first condition is necessary for consistency, (2) that the second condition is necessary, and (3) that both conditions together are sufficient.

**Part 1:** *If $D$ is consistent with $M$ then Condition 1 holds.*

Since $D$ is consistent with $M$, independence in $M$ is identical to that in $D$, so it is enough to show that two nodes are adjacent in $D$ iff there is no way to d-separate them.

A link between two adjacent nodes is a path which cannot be deactivated, thus if $\overline{ab}$ then there could not be any set $S$ s.t. $I(a, b|S) \in M$.

It remains to show that if there is no set $S$ s.t. $I(a, b|S) \in M$ then then $a$ and $b$ are adjacent. It suffices to consider $S = \{x \neq a, b : x$ is an ancestor of $a$ or $b\}$. Since, by assumption, $a$ and $b$ are not d-separated by any set, it must be the case that $I(a, b|S) \notin M$ thus there must be a path $\rho$ connecting $a$ and $b$ in $D$ which is active given $S$. Since $\rho$ is $S$-active, every head to head node on $\rho$ must be in or have a descendent in $S$. But by the definition of $S$, every node which has a descendant in $S$ must be in $S$ as well. Thus every head-to-head node on $\rho$ must be in $S$. Every other node on $\rho$ is an ancestor of $a$, $b$ or one of the head to head nodes of the path. Hence every node on $\rho$ must be in $S$ with the exception of $a$ and $b$. Thus every node of $\rho$, except $a$ and $b$, must be a head-to-head node. There are only three paths satisfying this condition: $a \rightarrow b$, $a \leftarrow b$ and $a \rightarrow c \leftarrow b$. However the last case is not possible because $c$ is in $S$ so it must be an ancestor of either $a$ or $b$ and thus it cannot be common child of both $a$ and $b$ as well or there would be a directed cycle. Hence $a$ and $b$ are adjacent.

**Part 2** *If $D$ is consistent with $M$ then Condition 2 holds.*

If $b$ is head-to-head in between $a$ and $c$ then the two link path cannot be de-activated by any set containing $b$. The rest of the only-if portion of condition 2 follows trivially from the definition of a vee structure.

To complete the proof of Part 2, let $\overline{abc}$ be a chain with $\neg\overline{ac}$. Furthermore, assume that for any set $S$, $I(a, b|S) \in M$ implies $b \notin S$. If $b$ were not head-to-head on the path $\overline{abc}$ then any set $S$ for which $I(a, c|S) \in M$ would necessarily contain $b$ in order to deactivate this path. Since $\neg\overline{ac}$, there must be a such an $S$, however by assumption for any such $S$, $b \notin S$. Thus $b$ must be head-to-head on the path $\overline{abc}$, hence it must be the case that $\overrightarrow{abc}$.

**Part 3** *If Conditions 1 and 2 hold then $D$ is consistent with $M$.*

If $M$ is dag isomorphic then there must exist a dag which is consistent with $M$, call it $D^*$. By Parts 1 and 2 above, $D$ and $D^*$ have the same skeletons and vee structures, so it is enough to prove Proposition A.2:

**Proposition A.2** *If any two dags, $D$ and $E$, have the same skeletons and vee structures then every active path in one dag corresponds to an active path in the other.*

Let $\rho$ be an $S$-active path in $D$ which is minimal in the following sense: if $k$ is the number of nodes in $\rho$,

$\rho_1$ is the first node and $\rho_k$ is the last node then (1) there cannot exist an $S$-active path between $\rho_1$ and $\rho_k$ with strictly fewer than $k$ nodes and (2) there cannot exist a different $S$-active path $\phi$ between $\rho_1$ and $\rho_k$ with exactly $k$ nodes such that for all $1 < i < k$, either $\phi_i = \rho_i$ or $\phi_1$ is a descendant of $\rho_i$.

Since $D$ and $E$ have the same links $\rho$ must be a path in $E$. It can be shown by induction on the number of head-to-head nodes that $\rho$ is $S$-active in $E$ as well. By definition, a single nodes will be considered as an active path. The remainder of the proof has three sub-parts: the first part proves that if $\rho$ contains no head-to-head nodes then it is $S$-active in $E$, the second part proves that if $\rho$ contains at least one head-to-head node $x = \rho_i$ then $\rho$ is $S$-active in $E$ iff $x$ is $S$-active in $E$, and the third part proves that $x$ is $S$-active in $E$.

Sub-Part 1:

If $\rho$ does not contain any head-to-head nodes in $D$ then it would be $S$-active in $E$ unless it contains a head-to-head node in $E$. It is enough to show that $\rho$ cannot have any head-to-head nodes in $E$. Suppose that some node $x = \rho_i$ were head-to-head in $E$ with parents $y = \rho_{i-1}$ and $z = \rho_{i+1}$, Figure 2 shows the possible configurations for $D$.

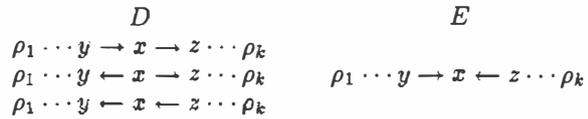

$$
\begin{array}{ll}
\qquad\qquad D & \qquad\qquad E \\
\rho_1 \cdots y \rightarrow x \rightarrow z \cdots \rho_k \\
\rho_1 \cdots y \leftarrow x \rightarrow z \cdots \rho_k & \rho_1 \cdots y \rightarrow x \leftarrow z \cdots \rho_k \\
\rho_1 \cdots y \leftarrow x \leftarrow z \cdots \rho_k
\end{array}
$$

Figure 2: $D$ has no head-to-head nodes, but $E$ does.

The parents of $\rho_i$ along $\rho$ in $E$ would be adjacent in both $D$ and $E$ since the two graphs share links and vee structures. But the sequence of nodes formed by removing $\rho_i$ from $\rho$ would be a path in $D$ since its parents would be adjacent. Moreover this path would be $S$-active since it could contain no head-to-head nodes (unless $D$ contained a directed loop). But this path would contradict Condition 1 of the minimality of $\rho$ in $D$. Therefore if $\rho$ contains no head-to-head nodes in $D$ then it is $S$-active in $E$.

Sub-Part 2:

Suppose that $\rho$ contains at least one head-to-head node $x = \rho_i$ in $D$ with parents $y = \rho_{i-1}$ and $z = \rho_{i+1}$ as shown in Figure 3. Let $\rho_{1,i-1}$ be the subpath of $\rho$ be-

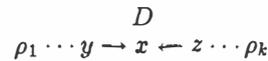

$$
\begin{array}{c}
D \\
\rho_1 \cdots y \rightarrow x \leftarrow z \cdots \rho_k
\end{array}
$$

Figure 3: $D$ has some head-to-head nodes.

tween $a$ and $y$ and $\rho_{i+1,k}$ be the subpath between $z$ and $b$. Note that $i - 1$ may equal 1 and/or $i + 1$ may equal $k$, in which case the corresponding subpath(s) would be a single node. Both $\rho_1$ and $\rho_2$ are minimal $S$-active paths of $D$ and both contain strictly fewer head-to-head nodes than $\rho$ thus by the inductive hypothesis, they are $S$-active in $E$. If $y$ and $z$ were adjacent in $D$ then since both nodes are both $S$-active in $D$ (they are parents of an $S$-active node) and neither is in $S$ (because neither is head-to-head on $\rho$ in $D$), it follows that the path formed by removing $x$ from $\rho$ would be $S$-active. This path which would contradict Condition 1 of the minimality of $\rho$.

Therefore $y$ and $z$ cannot be adjacent in either graph and must be common parents of $x$ in both. Since $x$ is head-to-head on $\rho$ in $E$ and both the subpaths $\rho_{1,i-1}$ and $\rho_{i+1,k}$ are $S$-active in $E$ it follows that $\rho$ would be $S$-active in $E$ iff $x$ were $S$-active in $E$.

Sub-Part 3:

Since $x$ is $S$-active in $D$ there exists a directed path in $D$ from $x$ to some node $w$ in $S$. Let $\phi$ be the shortest such path. It remains to show (by induction on the length $l$ of $\phi$) that $\phi$ is strictly directed from $x$ to $w$ in $E$. There are three cases, either $l = 0$, $l = 1$ or $l > 1$.

If $l = 0$ then $x = w$ and $x$ is trivially $S$-active in $E$.

If $l = 1$ then $\phi$ is a single link. Consider the parents, $y$ and $z$ of $x$. If they were both adjacent to $w$ as in

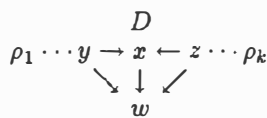

$$\rho_1 \cdots y \rightarrow x \leftarrow z \cdots \rho_k$$

Figure 4: A single link descendant path.

Figure 4 then they would be common parents of $w$ in $D$ (or there would be a directed loop in $D$). Thus the sequence of nodes $\phi'$ formed by replacing $x$ with $w$ in $\rho$ would be an $S$-active path in $D$. This path would contradict Condition 2 of the minimality of $\rho$, so at least one parent of $x$ must not be adjacent to $w$. Without loss of generality, assume $y$ is not adjacent to $w$. Since $y$ and $w$ are not parents of $x$ in $D$, they cannot both be parents of $x$ in $E$ as the two graphs share vee structures. Therefore $x$ must be a parent of $w$ in $E$ and $x$ would be $S$-active in $E$.

If $l > 1$ then $\phi$ contains at least two links. Consider the last two links of $\phi$, shown in Figure 5 where $u = \phi_l - 2$, $v = \phi_l - 1$. Note that $l - 2$ may equal $1$ in which case $x = u$. The initial subpath $\phi_{1,l-1}$ must be directed from $x$ to $v$ by induction. If $u$ were adjacent to $w$ then there would have been a shorter directed path from $x$ to $w$ in $D$, thus $u$ and $w$ are not adjacent and not parents of $v$ in $D$ so they cannot both be parents of $v$ in $E$. Therefore $v$ must be a parent of $w$ in $E$ and $\phi$ is $S$-active in $E$.    □

**Corollary 3.2** *Two dags are equivalent iff they share the same set of links and same set of vee structures.*

**Proof:** This result follows directly from the proof of the previous lemma.    □

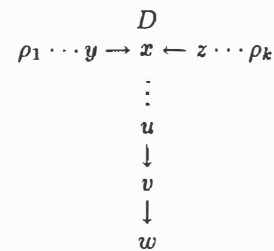

$$\rho_1 \cdots y \rightarrow x \leftarrow z \cdots \rho_k$$

$$\vdots$$

$$u$$

$$\downarrow$$

$$v$$

$$\downarrow$$

$$w$$

Figure 5: A multiple link descendant path.

**Lemma 3.3** *For any dag-isomorphic dependency model $M$ and any chain $\overline{abc}$,*
*if $\exists_S$ s.t. $I(a, c|S) \in M$ and $b \notin S$ then $\forall'_S\, I(a, c|S') \in M$ implies $b \notin S'$.*

**Proof:** Suppose $\overline{abc}$ and $\exists_S$ s.t. $I(a, c|S) \in M$ and $b \notin S$. In order for $S$ to d-separate $a$ and $c$, it must be the case that $a \rightarrow b \leftarrow c$ — if $b$ were not head-to-head then this two link path would be active given any set not containing $b$. Now since $b$ is head-to-head it must be the case that any set $S$ which contains $b$ will activate this two link path, hence for any $S$ if $I(a, b|S) \in M$ then $b \notin S$.    □

**Definition A.3** *A graphoid is a dependency model satisfying the following four axioms:*

| | | | |
|---|---|---|---|
| symmetry | $I(X, Y|Z)$ | $\Leftrightarrow$ | $I(Y, X|Z)$ |
| decomposition | $I(W, XY|Z)$ | $\Rightarrow$ | $I(W, Y|Z)$ |
| weak union | $I(W, XY|Z)$ | $\Rightarrow$ | $I(W, X|YZ)$ |
| contraction | $I(W, Y|Z) \wedge I(W, X|YZ)$ | | |
| | | $\Rightarrow$ | $I(W, XY|Z)$ |

**Lemma 5.1** *If $\exists_S$ s.t. $I(a, b|S) \in CL(L)$ then $\exists_{A,B,C}$ s.t. $I(aA, bB|C) \in L$*

**Proof:** This can be proven by induction on the derivation of $I(a, b|S)$. If the derivation has length $0$ then the lemma is trivial. If it is of length $k$ then $I(a, b|S)$ must follow from one of the rules. Each rule has an antecedent with $a$ separated from $b$ in a manner satisfying the inductive hypothesis. Thus since this antecedent must have a derivation of length $< k$ the lemma holds.    □